\documentclass[12pt,letterpaper]{article}

\usepackage[preprint,nonatbib]{nips_2018}

\usepackage[utf8]{inputenc} 
\usepackage[T1]{fontenc}    
\usepackage{hyperref}       
\usepackage{url}            
\usepackage{booktabs}       
\usepackage{amsfonts}       
\usepackage{nicefrac}       
\usepackage{microtype}      
\usepackage{graphicx}
\usepackage[flushleft]{threeparttable}
\usepackage{bm}
\usepackage[backend=bibtex,sorting=none,style=numeric-comp]{biblatex}
\DeclareGraphicsExtensions{.pdf,.jpeg,.png,.jpg}

\title{Deep Learning to Improve Breast Cancer Early Detection on Screening Mammography}

\author{
  Li~Shen \\
  Department of Neuroscience \\
  Icahn School of Medicine at Mount Sinai (ISMMS) \\
  New York, 10029, USA \\
  \texttt{li.shen@mssm.edu} \\
  \And
  Laurie~R.~Margolies \\
  Department of Diagnostic, Molecular, and Interventional Radiology \\
  ISMMS \\
  \And
  Joseph~H.~Rothstein \\
  Department of Genetics and Genomic Sciences \\
  Department of Population Health Science and Policy \\
  ISMMS \\
  \And
  Eugene~Fluder \\
  Department of Scientific Computing \\
  ISMMS \\
  \And
  Russell~B.~McBride \\
  Department of Pathology \\
  ISMMS \\
  \And
  Weiva~Sieh \\
  Department of Population Health Science and Policy \\
  Department of Genetics and Genomic Sciences \\
  ISMMS
}

\begin{document}

\maketitle

\begin{abstract}
The rapid development of deep learning, a family of machine learning techniques, has spurred much interest in its application to medical imaging problems. Here, we develop a deep learning algorithm that can accurately detect breast cancer on screening mammograms using an "end-to-end" training approach that efficiently leverages training datasets with either complete clinical annotation or only the cancer status (label) of the whole image. In this approach, lesion annotations are required only in the initial training stage, and subsequent stages require only image-level labels, eliminating the reliance on rarely available lesion annotations. Our all convolutional network method for classifying screening mammograms attained excellent performance in comparison with previous methods. On an independent test set of digitized film mammograms from Digital Database for Screening Mammography (DDSM), the best single model achieved a per-image AUC of 0.88, and four-model averaging improved the AUC to 0.91 (sensitivity: 86.1\%, specificity: 80.1\%). On a validation set of full-field digital mammography (FFDM) images from the INbreast database, the best single model achieved a per-image AUC of 0.95, and four-model averaging improved the AUC to 0.98 (sensitivity: 86.7\%, specificity: 96.1\%). We also demonstrate that a whole image classifier trained using our end-to-end approach on the DDSM digitized film mammograms can be transferred to INbreast FFDM images using only a subset of the INbreast data for fine-tuning and without further reliance on the availability of lesion annotations. These findings show that automatic deep learning methods can be readily trained to attain high accuracy on heterogeneous mammography platforms, and hold tremendous promise for improving clinical tools to reduce false positive and false negative screening mammography results. Code and model available at: \url{https://github.com/lishen/end2end-all-conv}
\end{abstract}

\section{Introduction}
The rapid advancement of machine learning and especially deep learning continues to fuel the medical imaging community's interest in applying these techniques to improve the accuracy of cancer screening. Breast cancer is the second leading cause of cancer deaths among U.S. women \cite{american_cancer_society_how_2018} and screening mammography has been found to reduce mortality \cite{oeffinger_breast_2015}. Despite the benefits, screening mammography is associated with a high risk of false positives as well as false negatives. According to a study conducted by the Breast Cancer Surveillance Consortium in 2009, the overall sensitivity of digital screening mammography in the U.S. is 84.4\% and the overall specificity is 90.8\% \cite{breast_cancer_surveillance_consortium_performance_2009}. To help radiologists improve the predictive accuracy of screening mammography, computer-assisted detection and diagnosis (CAD) software (reviewed in \cite{elter_cadx_2009}) have been developed and in clinical use since the 1990s. Unfortunately, data suggests that commercial CAD systems have not led to significant improvement in performance \cite{fenton_influence_2007,cole_impact_2014,lehman_diagnostic_2015} and progress has stagnated in the past decade. With the remarkable success of deep learning in visual object recognition and detection, and many other domains \cite{lecun_deep_2015}, there is much interest in developing deep learning tools to assist radiologists and improve the accuracy of screening mammography \cite{aboutalib_deep_2018,kim_applying_2018,hamidinekoo_deep_2018,burt_deep_2018}.

Early detection of subclinical breast cancer on screening mammography is challenging as an image classification task because the tumors themselves occupy only a small portion of the image of the entire breast. For example, a full-field digital mammography (FFDM) image is typically $4000 \times 3000$ pixels while a cancerous region of interest (ROI) can be as small as $100 \times 100$ pixels. If ROI annotations were widely available in mammography databases then established object detection and classification methods such as the region-based convolutional neural network (R-CNN) \cite{girshick_rich_2014} and its variants \cite{girshick_fast_2015,ren_faster_2015,dai_r-fcn_2016} could be readily applied. However, approaches that require ROI annotations \cite{jamieson_breast_2012,arevalo_convolutional_2015,carneiro_unregistered_2015,dhungel_automated_2015,ertosun_probabilistic_2015,akselrod-ballin_region_2016,arevalo_representation_2016,levy_breast_2016,dhungel_automated_2016,becker_deep_2017,ribli_detecting_2017} often cannot be transferred to large mammography databases that lack ROI annotations, which are laborious and costly to assemble. Indeed, few public mammography databases are annotated \cite{moreira_inbreast_2012}. Yet, deep learning requires large training datasets to be most effective. Thus, it is essential to leverage both the few fully annotated datasets, as well as larger datasets labeled with only the cancer status of each image to improve the accuracy of breast cancer classification algorithms.

Pre-training is a promising method to address the training problem. For example, Hinton et al. \cite{hinton_fast_2006} used layer-wise pre-training to initialize the weight parameters of a deep belief net (DBN) with three hidden layers and then fine-tuned it for classification. They found that pre-training improved the training speed as well as the accuracy of handwritten digit recognition. Another popular training method is to first train a deep learning model on a large database such as the ImageNet \cite{russakovsky_imagenet_2015} and then fine-tune the model for another task. Although the specific task may not be related to the initial training dataset, the model's weight parameters are already initialized for recognizing primitive features, such as edges, corners and textures, which can be readily used for a different task. This often saves training time and improves the model's performance.

In this study, we propose an "end-to-end" approach in which a model to classify local image patches is pre-trained using a fully annotated dataset with ROI information. The patch classifier's weight parameters are then used to initialize the weight parameters of the whole image classifier, which can be further fine-tuned using datasets without ROI annotations (see discussion in \cite{shen_breast_2017}). We used a large public film-based mammography database---with thousands of images---to develop the patch and whole image classifiers, and then transferred the whole image classifiers to a public FFDM database---with hundreds of images. We evaluated various network designs for constructing the patch and whole image classifiers to attain the best performance. The pipeline required to build a whole image classifier is presented here, as well as the pros and cons of different training strategies.

\section{Methods}
\subsection{Converting a classifier from recognizing patches to whole images}
\begin{figure}
  \centering
  \includegraphics[width=\linewidth]{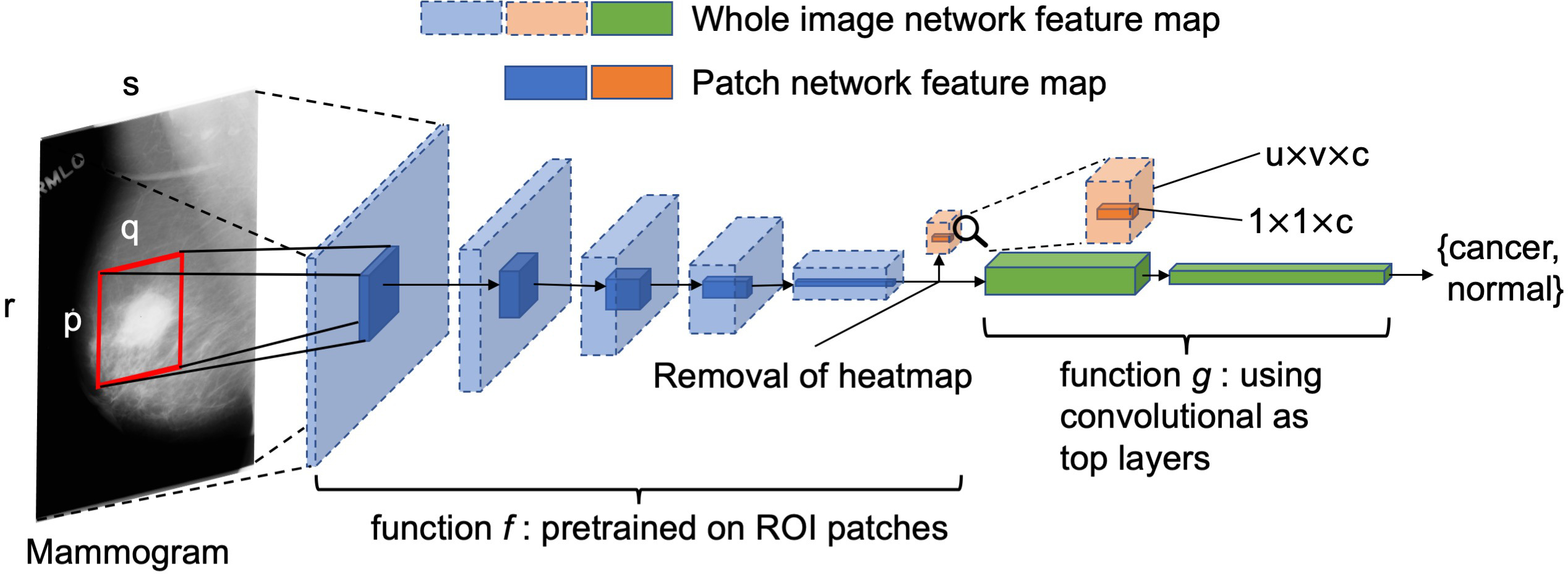}
  \caption{Converting a patch classifier to an end-to-end trainable whole image classifier using an all convolutional design where we considered removing the heatmap to improve information flow and convolutional layers as top layers; the magnifying glass shows an enlarged version of the heatmap.}
  \label{fig:patch2image}
\end{figure}

To perform classification or segmentation on large complex images, a common strategy involves the use of a classifier in sliding window fashion to recognize local patches on an image to generate a grid of probabilistic outputs. This is followed by another process to summarize the patch classifier's outputs to give the final classification or segmentation result. Such methods have been used to detect metastatic breast cancer using whole slide images of sentinel lymph node biopsies \cite{wang_deep_2016} and to segment neuronal membranes in microscopic images \cite{ciresan_deep_2012}. However, this strategy requires two steps that each needs to be optimized separately. Here, we propose a method to combine the two steps into a single step for training on the whole images (Fig.~\ref{fig:patch2image}). Assume we have an input patch $X\in \rm I\!R^{p\times q}$ and a patch classifier which is a function $f$ so that $f(X)\in \rm I\!R^{c}$, where $c$ is the number of categories that the patch classifier recognizes; the function's output satisfies $f(X)_{i} \in [0, 1]$ and $\Sigma_{i=1}^{c}f(X)_{i}=1$. $c$ is typically a small integer, e.g. $c=5$ represents the benign-calcification, malignant-calcification, benign-mass, malignant-mass and background classes of a patch from a mammogram. Assume the input patch is extracted from an image $M\in \rm I\!R^{r\times s}$ where $p \ll r$ and $q \ll s$. If the function $f$ represents a convolutional neural network (CNN), then $f$ can be applied to $M$ without changing the network parameters so that $f(M)\in \rm I\!R^{u \times v \times c}$, where $u > 1$ and $v > 1$ depend on the image size and the stride of the patch classifier. This is possible because of the weight sharing and locality properties of a CNN. If the function $f$ represents a different class of neural networks, such as the multilayer perceptron (MLP), then this becomes infeasible since a MLP requires the input to be fixed. Therefore, after changing the input from $X$ to $M$, we have a $u \times v$ grid of probabilistic outputs of $c$ classes (referred to as "heatmap") instead of a single output of $c$ classes. Hence the heatmap has a size of $u \times v \times c$. More layers can then be added on top of the heatmap to transform the outputs and connect with the final classification output of the image. Adding a convolutional layer on top of the patch classifier's outputs turns the entire patch classifier into a filter and enlarges its receptive field. For example, if the patch classifier has a receptive field of $224 \times 224$ with a stride$=32$, adding a $3 \times 3$ convolutional layer on top of it increases each side of the receptive field to $224+(3-1) \times 32=288$. Thus, the top layers effectively use the patch classifier to "scan" the whole image, looking for cues of cancerous lesions and extracting higher level features that can finally be used for whole image classification. Using function $g$ to represent the top layers, the whole image classification function can be written as $h(M) = g(f(M)) \in \rm I\!R^{d}$, where $d$ is the number of classes of the whole image. Typically, $d = 2$ represents the two classes we want to predict: benign (or normal) and malignant. 

The function $h$ accepts whole images as input and produces labels at the whole image level. Therefore, it is end-to-end trainable, providing two advantages against the two-step approach: First, the entire network can be jointly trained, avoiding sub-optimal solutions from each step; Second, the trained network can be transferred to another dataset without explicit reliance on ROI annotations. Large mammography databases with ROI annotations are rare and expensive to obtain. The largest public database with ROI annotations for digitized film mammograms – DDSM \cite{heath_digital_2001} – contains several thousand images with pixel-level annotations, which can be exploited to train a patch classifier $f$. Once the patch classifier is converted into a whole image classifier $h$, it can be fine-tuned on other databases using only image-level labels. This approach allows us to significantly reduce the requirement for ROI annotations, and has many applications in medical imaging in addition to mammographic breast cancer classification. Practically, variable input size is a feature that is supported by most major deep learning frameworks \cite{chollet_keras_2015,abadi_tensorflow_2015,jia_caffe_2014,chen_mxnet_2015}, making it easy to implement. 

\subsection{Network design}
A modern CNN is typically constructed by stacking convolutional layers on top of the input, followed by one or more fully connected (FC) layers to join with the classification output. Max pooling layers are often used amid convolutional layers to achieve translational invariance and to reduce feature map size. In this study, two popular CNN structures are compared: the VGG network \cite{simonyan_very_2014} and the residual network (Resnet) \cite{he_deep_2015}. Consecutive network layers can be naturally grouped into "blocks" so that the feature map size is reduced (typically by a factor of 2) either at the beginning or at the end of a block but stays the same elsewhere in the block. For example, a "VGG block" is a stack of several $3 \times 3$ convolutional layers with the same depth followed by a $2 \times 2$ max pooling layer that reduces the feature map size by a factor of 2. Although other filter sizes can be used, $3 \times 3$ convolution and $2 \times 2$ max pooling seem to be the most popular choices and are used throughout this study unless otherwise stated. Therefore, a VGG block can be represented by the pattern of $N \times K$, where $N$ represents the depth of each convolutional layer and $K$ represents the number of convolutional layers. A "Resnet block" uses stride$=2$ in the first convolutional layer instead of $2 \times 2$ max pooling to reduce feature map size at the beginning of the block, followed by the stacking of several convolutional layers. We use the "bottleneck design" \cite{he_deep_2015} which consists of repeated units of three convolutional layers that have filter sizes of $1 \times 1$, $3 \times 3$ and $1 \times 1$, respectively. A key feature of the Resnet block is that a shortcut is made between the two ends of each unit so that the features are directly carried over and therefore each unit can focus on learning the "residual" information \cite{he_deep_2015}. Batch normalization (BN) is used in every convolutional layer in the Resnet, which is known to speedup convergence and also has a regularization effect \cite{ioffe_batch_2015}. A Resnet block can be represented by the pattern of $[L-M-N] \times K$, where $L$, $M$ and $N$ represent the depths of the three convolutional layers in a unit and $K$ represents the number of units. Here, the 16-layer VGG network (VGG16) and the 50-layer Resnet (Resnet50) are used as patch classifiers. In the original design of the VGG16 \cite{simonyan_very_2014}, it consists of five VGG blocks followed by two FC layers. To be consistent with the Resnet50, we replace the two FC layers with a global average pooling layer which calculates the average activation of each feature map for the output of the last VGG block. For example, if the output of the last VGG block has a size of $7 \times 7 \times 512$ (height $\times$ width $\times$ channel), after the global average pooling layer the output becomes $512$. This output is then connected to the classification output with a FC layer.

A straightforward approach to construct a whole image classifier from a patch classifier involves flattening the heatmap and connect it to the image's classification output using FC layers. To increase the model's translational invariance to the patch classifier's output, a max pooling layer can be used after the heatmap. Further, a shortcut can be made between the heatmap and the output to make the training easier. The heatmap is directly from the patch classifier's output which uses the softmax activation:
\begin{equation}
f(\bm{z})_{j} = \frac{e^{z_{j}}}{\Sigma_{i=1}^{c}e^{z_{i}}}\textrm{ for }j = 1, ..., c
\end{equation}
However, the softmax activation diminishes gradients for large inputs, which may impede gradient flow when it is used in an intermediate layer. Therefore, the rectified linear units (ReLU) can be used instead:
\begin{equation}
f(\bm{z})_{j} = max(0, z_{j})\textrm{ for }j = 1, ..., c
\end{equation}
In the following, when we refer to the heatmap in a whole image classifier, the activation is always assumed to be ReLU unless otherwise stated.

We further propose to use convolutional layers as top layers, which preserve spatial information. Two blocks of convolutional layers (VGG or Resnet) can be added on top of the patch classifier layers, followed by a global average pooling layer and then the image's classification output (Fig.~\ref{fig:patch2image}). Therefore, this design creates an "all convolutional" network for whole image classification. As Fig.~\ref{fig:patch2image} shows, the heatmap creates a bottleneck-like structure between the patch classifier layers and the top layers, which may cause information loss in the whole image classification. We remove the heatmap entirely from the whole image classifier to allow the top layers to fully utilize the features extracted from the patch classifier and compare the two choices in the following.

\section{Results}
\label{results}
\subsection{Developing patch and whole image classifiers on DDSM}
\subsubsection{Setup and processing of the dataset}
The DDSM \cite{heath_digital_2001} contains digitized film mammograms in a lossless-JPEG format that is obsolete. We used a later version of the database called CBIS-DDSM \cite{lee_curated_2016} which contains images that are converted into the standard DICOM format. The dataset which consisted of 2478 mammography images from 1249 women was downloaded from the CBIS-DDSM website and included both craniocaudal (CC) and mediolateral oblique (MLO) views for most of the exams. Each view was treated as a separate image in this study due to the sample size limit. The purpose of this study was to predict the malignant vs. benign (or normal) status of each image. We performed an 85-15 split on the patient-level data to create independent training and test sets. In the training set, we further isolated 10\% of the patients to create an independent validation set. The splits were done in a stratified fashion to maintain the same proportion of cancer cases in the training, validation and test sets. The total numbers of images in the training, validation and testing sets were: 1903, 199 and 376, respectively. 

The DDSM database contains the pixel-level annotations for the ROIs and their pathologically confirmed labels: benign or malignant. It further labels each ROI as a calcification or mass. Most mammograms contain only one ROI. All mammograms were converted into PNG format and down-sized to $1152 \times 896$. Two patch image sets were created by sampling image patches from ROIs and background regions. All patches had the same size of $224 \times 224$. The first dataset (S1) was comprised of sets of patches in which one is centered on the ROI and one is a random background patch from the same image. The second dataset (S10) was derived from 10 sampled patches around each ROI with a minimum overlapping ratio of 0.9 and the same number of background patches from the same image. All patches were classified into one of the five categories: background, malignant mass, benign mass, malignant calcification and benign calcification.

\subsubsection{Network training}
Training a whole image classifier was achieved in two steps. The first step was to train a patch classifier. We compared the networks with pre-trained weights using the ImageNet \cite{russakovsky_imagenet_2015} database to those with randomly initialized weights. For a pre-trained network, the bottom layers represent primitive features that tend to be preserved across different tasks, whereas the top layers represent higher-order features that are more related to specific tasks and require further training. Using the same learning rate for all layers may destroy the features that are already learned in the bottom layers. To prevent this, a 3-stage training strategy was proposed which freezes the parameter learning for all but the final layer and progressively unfreezes parameter learning from top to bottom, while simultaneously decreasing the learning rate. The 3-stage training strategy on the S10 patch set was as follows:
\begin{enumerate}
  \item Set learning rate to 1e-3 and train the last layer for 3 epochs.
  \item Set learning rate to 1e-4, unfreeze the top layers and train for 10 epochs, where the top layer number is set to 46 for Resnet50 and 11 for VGG16.
  \item Set learning rate to 1e-5, unfreeze all layers and train for 37 epochs for a total of 50 epochs.
\end{enumerate}
In the above, an epoch was defined as a sweep through the training set. For the S1 patch set, the total number of epochs was increased to 200 because it is much smaller than the S10 patch set. For randomly initialized networks a constant learning rate of 1e-3 was used. Adam \cite{kingma_adam_2014} was used as the optimizer and the batch size was set to be 32. The sample weights were adjusted within each batch to keep the five classes balanced.

The second step was to train a whole image classifier converted from the patch classifier (Fig.~\ref{fig:patch2image}). Similarly, a 2-stage training strategy was used to first train the newly added top layers (i.e. function $g$) and then train all layers (i.e. function $h$) with a reduced learning rate, which was as follows:
\begin{enumerate}
  \item Set learning rate to 1e-4, weight decay to 0.001 and train the newly added top layers for 30 epochs. 
  \item Set learning rate to 1e-5, weight decay to 0.01 and train all layers for 20 epochs for a total of 50 epochs.
\end{enumerate}
If the above 50 epochs did not lead to convergence, the training was resumed with up to 200 additional epochs. Due to GPU memory limit, a batch size of 2 was used.

The average gray scale value of the whole image training set was subtracted from both patch and whole image sets in training. No other preprocessing was applied. Data augmentation was used on-the-fly by doing the following random transformations: horizontal and vertical flips, rotation in [-25, 25] degrees, shear in [-0.2, 0.2] radians, zoom in [0.8, 1.2] ratio and channel shift in [-20, 20] pixel values. 

\subsubsection{Development of patch classifiers}
The accuracy of Resnet50 and VGG16 on the test set is reported in Table~\ref{tab:patch_acc}. The S10 set is more difficult to classify than the S1 set because the former contains the ROIs' neighboring and background regions that are challenging to distinguish. On the S1 set, both randomly initialized and pre-trained Resnet50 classifiers achieved similar accuracy but the pre-trained network converged after half as many epochs as the randomly initialized one. On the S10 set, the pre-trained Resnet50 outperformed the randomly initialized one by a large margin, achieving an accuracy of 0.89. These results showed that pre-training can greatly help network convergence and performance. Therefore, pre-trained networks were used for the rest of the study. The pre-trained VGG16 achieved an accuracy of 0.84 on the S10 set, falling short of the pre-trained Resnet50's performance.

\begin{table}
\centering
\caption{Test accuracy of the patch classifiers using the Resnet50 and VGG16. \#Epochs indicates the epoch when the best validation accuracy has been reached.}
\label{tab:patch_acc}
\begin{tabular}{@{}lllll@{}}
\toprule
\textbf{Model} & \textbf{Pretrained} & \textbf{Patch set} & \textbf{Accuracy} & \textbf{\#Epochs} \\ \midrule
Resnet50 & N & S1 & 0.97 & 198 \\
Resnet50 & Y & S1 & 0.99 & 99 \\
Resnet50 & N & S10 & 0.63 & 24 \\
Resnet50 & Y & S10 & 0.89 & 39 \\
VGG16 & Y & S10 & 0.84 & 25 \\ \bottomrule
\end{tabular}
\end{table}

\subsubsection{Converting patch to whole image classifiers}
Using pre-trained Resnet50 and VGG16 as patch classifiers, we tested several different configurations for the top layers. For the following configurations, we removed the heatmap and added two Resnet or VGG blocks on top of the patch classifier layers, followed by a global average pooling layer and the classification output. The models were evaluated by per-image AUCs on the test set. 

\textbf{Resnet-based networks}
\label{resnet_auc}
To evaluate whether the patch classifiers trained on the S1 and S10 sets are equally useful for whole image classification, the Resnet50 patch classifiers were used, followed by two Resnet blocks of the same configuration of $[512-512-2048] \times 1$ (Table~\ref{tab:resnet_auc}). The whole image classifier based on the S10 set (AUC=0.85) performed much better than the one based on the S1 set (AUC=0.63). The S10 set contains more information about the ROIs as well as their adjacent regions and background regions than the S1 set, which can be important for whole image classification. For the rest of the study, only patch classifiers trained on the S10 set were used. Varying the configuration by using two Resnet blocks of the same configuration of $[512-512-1024] \times 2$ yielded the same AUC of 0.85 while reducing the depths of the two Resnet blocks decreased the AUC by 0.01-0.02 (Table~\ref{tab:resnet_auc}). This result showed that the depths for the Resnet blocks were relatively uncorrelated with the performance of the whole image classifiers.

\begin{table}
\centering
\caption{Per-image test AUC scores of the whole image classifiers using the Resnet50 as patch classifiers. \#Epochs indicates the epoch when the best validation score has been reached. Bold style corresponds to the best performing models.}
\label{tab:resnet_auc}
\begin{tabular}{@{}llllll@{}}
\toprule
\textbf{Patch set} & \textbf{Block1} & \textbf{Block2} & \textbf{AUC} & \textbf{A-AUC} & \textbf{\#Epochs} \\ \midrule
S1 & {[}512-512-2048{]} x 1 & {[}512-512-2048{]} x 1 & 0.63 & NA & 35 \\
S10 & {[}512-512-2048{]} x 1 & {[}512-512-2048{]} x 1 & 0.85 & 0.87 & 20 \\
\textbf{S10} & \textbf{{[}512-512-1024{]} x 2} & \textbf{{[}512-512-1024{]} x 2} & \textbf{0.85} & \textbf{0.87} & \textbf{34} \\
S10 & {[}256-256-256{]} x 1 & {[}128-128-128{]} x 1 & 0.84 & 0.85 & 25 \\
S10 & {[}256-256-256{]} x 3 & {[}128-128-128{]} x 3 & 0.83 & 0.84 & 17 \\
S10 & {[}256-256-512{]} x 3 & {[}128-128-256{]} x 3 & 0.84 & 0.87 & 48 \\
\textbf{S10} & \textbf{256 x 1} & \textbf{128 x 1} & \textbf{0.87} & \textbf{0.88} & \textbf{36} \\ \midrule
\multicolumn{6}{c}{\textbf{Insert heatmap between patch classifier and top layers}} \\ \midrule
S10 & {[}512-512-1024{]} x 2 & {[}512-512-1024{]} x 2 & 0.80 & NA & 47 \\
S10 & {[}64-64-256{]} x 2 & {[}128-128-512{]} x 2 & 0.81 & NA & 41 \\ \midrule
\multicolumn{6}{c}{\textbf{Add heatmap and FC layers on top (on S10)}} \\ \midrule
\textbf{Pool size} & \textbf{FC1} & \textbf{FC2} & \textbf{} & \textbf{} & \textbf{} \\
5x5 & 64 & 32 & 0.73 & NA & 28 \\
2x2 & 512 & 256 & 0.72 & NA & 47 \\
1x1 & 2048 & 1024 & 0.65 & NA & 43 \\ \bottomrule
\end{tabular}
\end{table}

\begin{figure}
  \centering
  \includegraphics[width=0.7\linewidth]{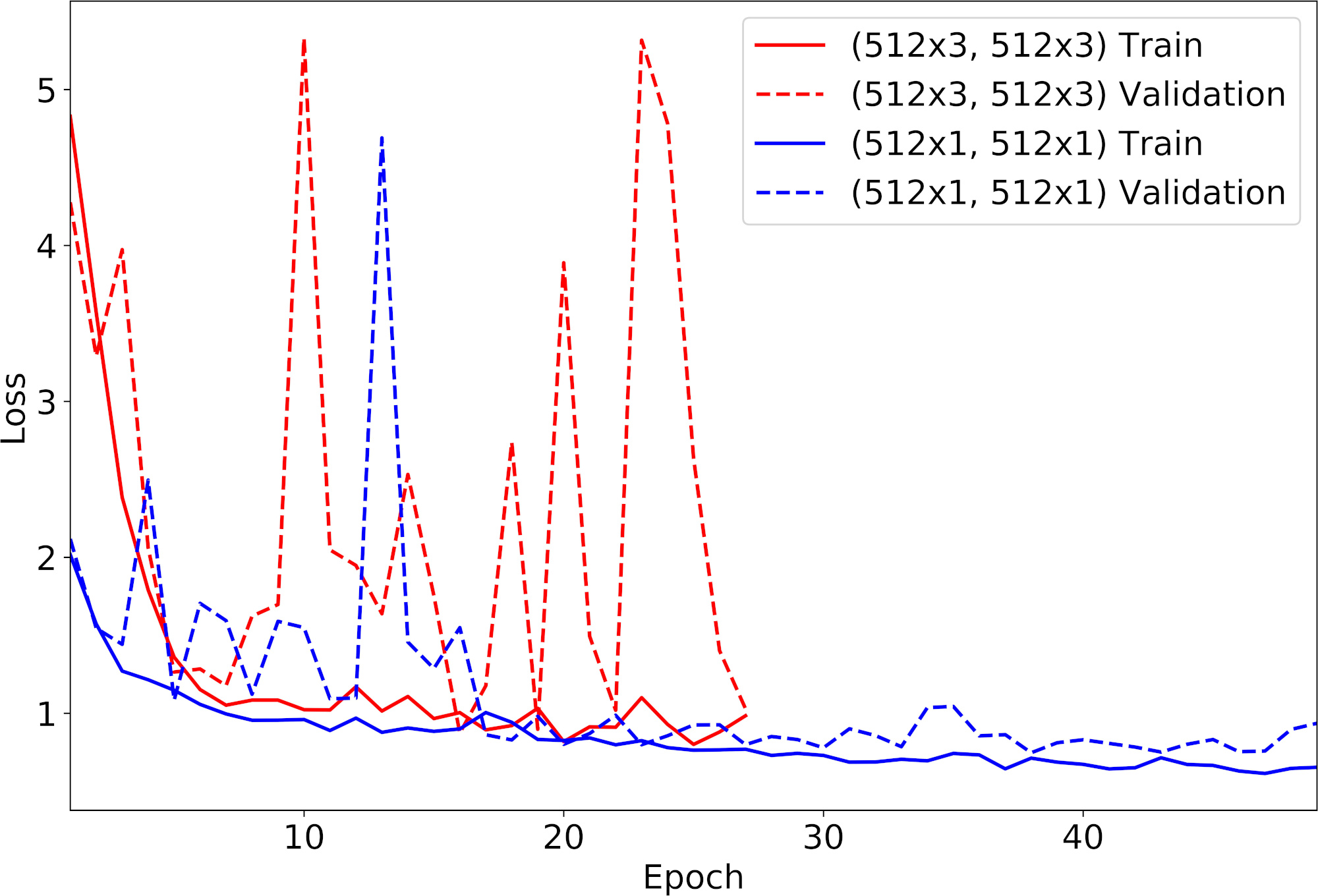}
  \caption{Train and validation loss curves of two VGG structures.}
  \label{fig:vgg_loss}
\end{figure}

\textbf{VGG-based networks}
\label{vgg_auc}
We then tested whole image classifiers using VGG16 as the patch classifier and VGG blocks as the top layers. BN was used for the VGG blocks on the top but not for the VGG16 patch classifier because it is a pre-trained network which cannot be modified. The VGG-based whole image classifiers performed similarly to the Resnet-based ones but took longer to achieve the same performance (Table~\ref{tab:vgg_auc}). Using more convolutional layers in VGG blocks led to poorer performance: using two VGG blocks of $512 \times 1$ (AUC=0.83) performed better than two VGG blocks of $512 \times 3$ (AUC=0.81); using two VGG blocks of $256 \times 1$ and $128 \times 1$ (AUC=0.85) performed better than two VGG blocks of $256 \times 3$ and $128 \times 3$ (AUC=0.83). On the contrary to the Resnet, reducing the depths of the two VGG blocks from $512$ and $512$ to $256$ and $128$ increased the AUC. Reducing the depths further to $128$ and $64$ still yielded an AUC of 0.84. Fig.~\ref{fig:vgg_loss} illustrates that the classifier with the two VGG blocks of $512 \times 3$ suffered from overfitting: as the training loss decreased, the validation loss fluctuated wildly, while the classifier with the two VGG blocks of $512 \times 1$ showed smoother loss curves. Therefore, controlling model complexity (\#layers and depths) is important for achieving good performance with the VGG-based networks.

\begin{table}
\centering
\caption{Per-image test AUC scores of the whole image classifiers using the VGG16 as patch classifiers. \#Epochs indicates the epoch when the best validation score has been reached. Bold style corresponds to the best performing models.}
\label{tab:vgg_auc}
\begin{tabular}{@{}llllll@{}}
\toprule
\textbf{Patch set} & \textbf{Block1} & \textbf{Block2} & \textbf{AUC} & \textbf{A-AUC} & \textbf{\#Epochs} \\ \midrule
S10 & 512 x 3 & 512 x 3 & 0.81 & 0.82 & 91 \\
S10 & 512 x 1 & 512 x 1 & 0.83 & 0.86 & 98 \\
S10 & 256 x 3 & 128 x 3 & 0.83 & 0.85 & 51 \\
\textbf{S10} & \textbf{256 x 1} & \textbf{128 x 1} & \textbf{0.85} & \textbf{0.86} & \textbf{61} \\
S10 & 128 x 1 & 64 x 1 & 0.84 & 0.85 & 142 \\
\textbf{S10} & \textbf{{[}512-512-1024{]} x 2} & \textbf{{[}512-512-1024{]} x 2} & \textbf{0.85} & \textbf{0.88} & \textbf{165} \\ \midrule
\multicolumn{6}{c}{\textbf{Add heatmap and FC layers on top (on S10)}} \\ \midrule
\textbf{Pool size} & \textbf{FC1} & \textbf{FC2} & \textbf{} & \textbf{} & \textbf{} \\
5x5 & 64 & 32 & 0.71 & NA & 26 \\
2x2 & 512 & 256 & 0.68 & NA & 27 \\
1x1 & 2048 & 1024 & 0.69 & NA & 50 \\ \bottomrule
\end{tabular}
\end{table}

\textbf{Hybrid networks}
\label{hybrid_auc}
We also created two hybrid networks by adding the best VGG top layers - two VGG blocks of $256 \times 1$ and $128 \times 1$ - on top of the Resnet50 patch classifier; and the best Resnet top layers - two Resnet blocks of the same configuration of $[512-512-1024] \times 2$ on top of the VGG16 patch classifier. The two hybrid networks achieved AUC of 0.87 and 0.85, respectively; they are among the best performing models (Tables \ref{tab:resnet_auc} \& \ref{tab:vgg_auc}).

\begin{figure}
  \centering
  \includegraphics[width=\linewidth]{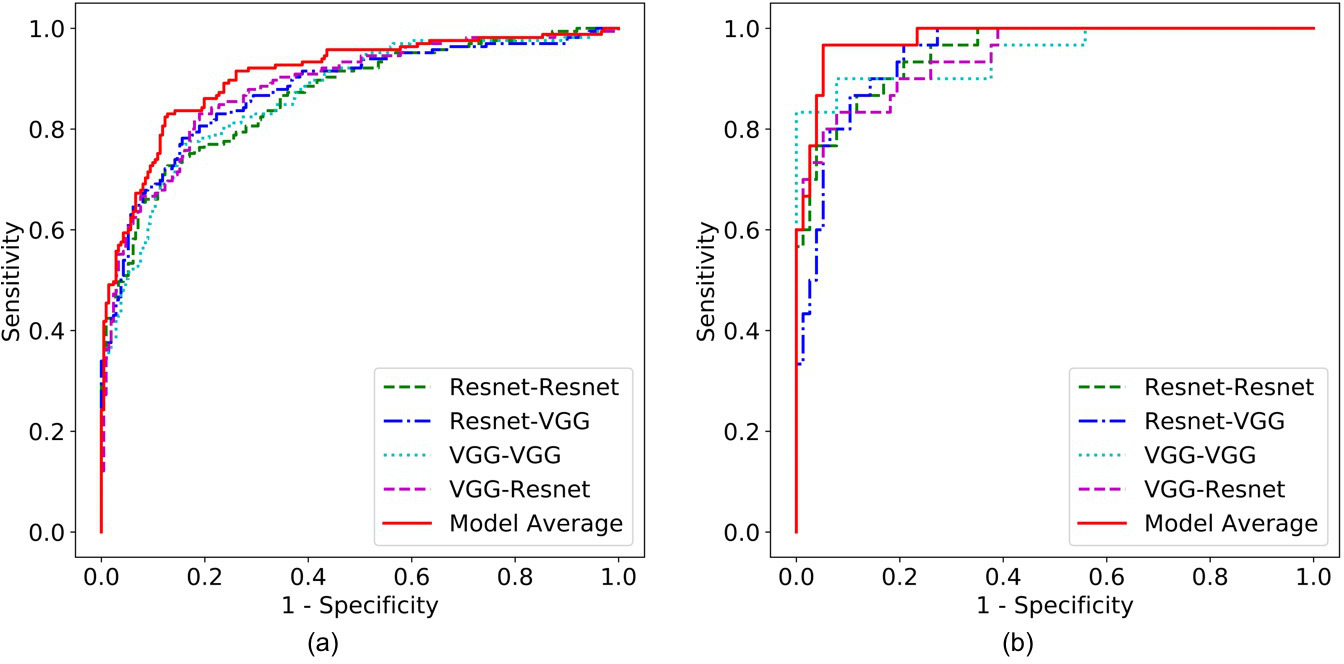}
  \caption{ROC curves of the four best performing models and their model ensemble on (a) The DDSM test set and (b) The INbreast validation set.}
  \label{fig:roc_curve}
\end{figure}

\textbf{Augmented prediction and model averaging}
\label{augmented_ensemble}
Augmented prediction was implemented by horizontally and vertically flipping an image to obtain four images and taking an average of the four images' scores. This technique increased the AUC (referred to as A-AUC) for each model by 0.01-0.03 (Tables \ref{tab:resnet_auc} \& \ref{tab:vgg_auc}). The four best performing models were chosen to make a model ensemble by taking an average of their augmented prediction scores: two models used Resnet50 and VGG16 as patch classifiers and Resnet and VGG blocks as top layers, respectively (referred to as Resnet-Resnet and VGG-VGG); and two hybrid models (referred to as Resnet-VGG and VGG-Resnet). Fig.~\ref{fig:roc_curve}a shows the Receiver Operating Characteristic (ROC) curves of the four best models and their model ensemble. The model ensemble yielded an AUC of 0.91, which corresponded to sensitivity of 86.1\% and specificity of 80.1\%.

\textbf{Using max-pooling, shortcut and FC layers}
\label{pool_fc_auc}
We tested the alternative design by using the heatmap followed by a max-pooling and two FC layers, including a shortcut between the heatmap and the classification output. The Resnet50 and VGG16 patch classifiers were used. The FC layer sizes were chosen to gradually reduce the layer outputs. When the pooling size increased from $1 \times 1$ (i.e. no pooling) to $5 \times 5$, the AUC also moderately increased (Tables \ref{tab:resnet_auc} \& \ref{tab:vgg_auc}) but even the best performing model achieved only AUC of 0.73, falling short of the all convolutional models.

\textbf{Using the heatmap in all convolutional networks}
\label{heatmap_auc}
To test our hypothesis about the heatmap being an information bottleneck in the all convolutional networks, we added the heatmap back into the Resnet-based whole image classifier with two $[512-512-1024] \times 2$ blocks as top layers. This model achieved an AUC of 0.80 (Table~\ref{tab:resnet_auc}), which was lower than that of the same classifier without the heatmap. To exclude the possibility of overfitting due to the small depth of the heatmap, another model with reduced complexity using two Resnet blocks of $[64-64-256] \times 2$ and $[128-128-512] \times 2$ was tested, which achieved an AUC of 0.81. We conclude that removing the heatmap is beneficial to the all convolutional networks.

\textbf{The two-step approach}
\label{two_step_auc}
Finally, for comparison we tested a previously reported approach \cite{shen_breast_2017,wang_deep_2016} that set a cutoff to binarize the heatmap; extracted regional features (such as area, major axis length and mean intensity) from the binary heatmap; and trained a random forest classifier (\#trees=500, max depth=9, min samples split=300) on the regional features. The Resnet50 patch classifier was used and the softmax activation was used in the heatmap to obtain the probabilities for the five classes. Four cutoffs---0.3, 0.5, 0.7 and 0.9 were used to binarize the heatmap and the regional features were combined. This approach achieved an AUC of 0.73, falling short of the all convolutional models.

\subsection{Transfer learning for whole image classification on INbreast}
\label{inbreast}
\subsubsection{Setup and processing of the dataset}
\begin{figure}
  \centering
  \includegraphics[width=0.7\linewidth]{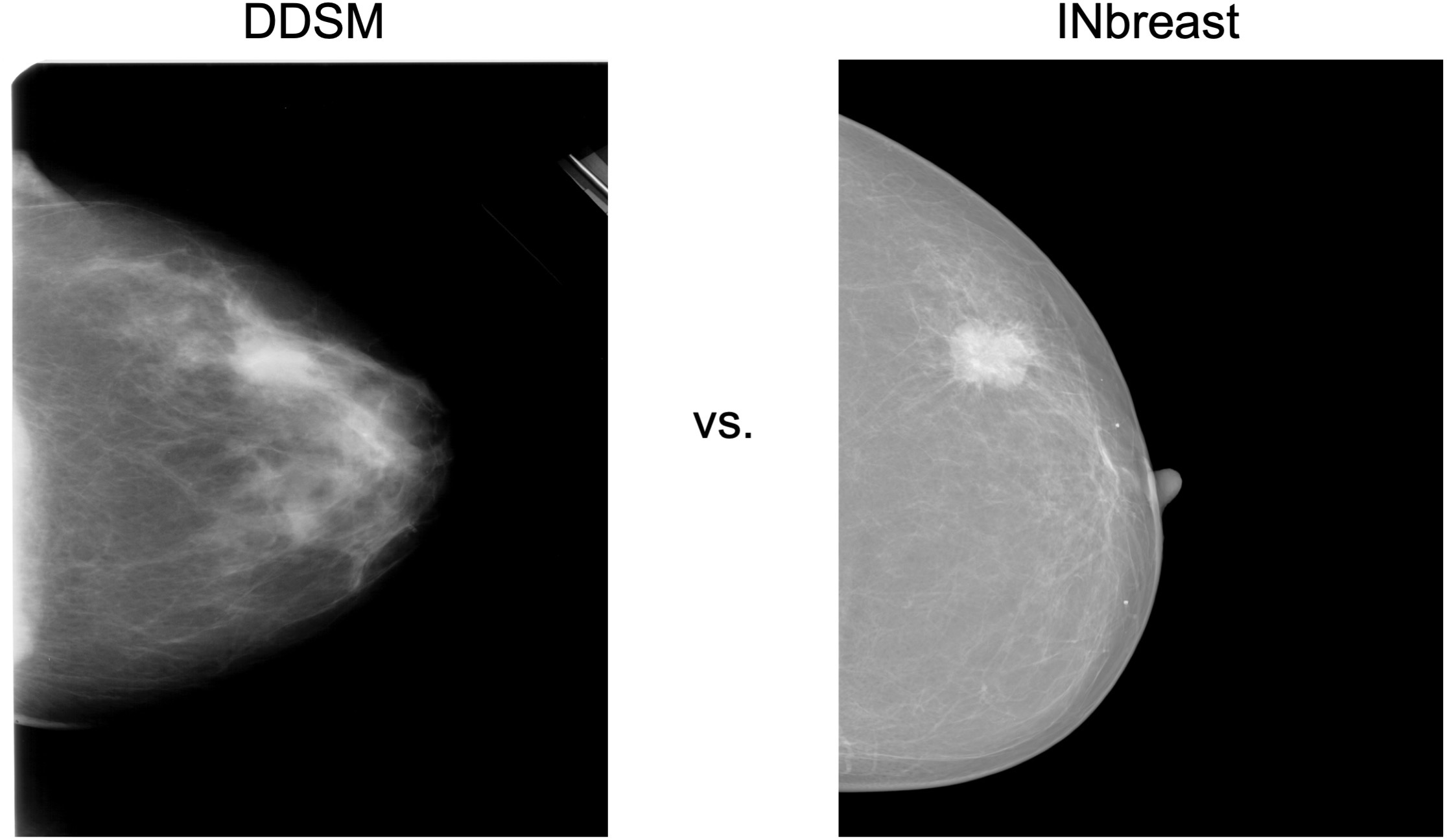}
  \caption{Comparison of representative mammograms from DDSM and INbreast.}
  \label{fig:ddsm_vs_inbr}
\end{figure}

The INbreast \cite{moreira_inbreast_2012} dataset is a more recent public database for mammograms that contains FFDM images as opposed to digitized film images. These images have different intensity profiles from the DDSM images, which can be visually confirmed by looking at two example images from the two databases (Fig.~\ref{fig:ddsm_vs_inbr}). Therefore, INbreast provides an excellent opportunity to test the transferability of a whole image classifier onto an independent dataset. The INbreast database contains 115 patients and 410 mammograms including both CC and MLO views. We analyzed each view separately due to sample size limit. The INbreast database includes radiologists' BI-RADS \cite{dorsi_acr_2013} assessments which are defined as follows: category 0, exam is not conclusive; category 1, no findings; category 2, benign findings; category 3, probably benign findings; category 4, suspicious findings; category 5, a high probability of malignancy; and category 6, proved cancer. Because the database lacks reliable pathological confirmation of malignancy, we assigned all images with BI-RADS 1 and 2 as negative; BI-RADS 4, 5 and 6 as positive; and excluded 12 patients and 23 images with BI-RADS 3 since this assessment is typically not given at screening. We split the dataset 70-30 into training and validation sets at the patient-level in a stratified fashion. The total numbers of images in the training and validation sets were 280 from 72 women and 107 from 31 women, respectively. We used the same processing steps on the INbreast images as for the DDSM images.

\subsubsection{Effectiveness and efficiency of transfer learning}
Although the INbreast database contains ROI annotations, they were ignored to test the transferability of the whole image classifier. The four best performing models (See Tables \ref{tab:resnet_auc} \& \ref{tab:vgg_auc}) were directly fine-tuned on the training set and evaluated by computing per-image AUCs on the validation set. Adam \cite{kingma_adam_2014} was used as the optimizer and the learning rate was set at 1e-5. The number of epochs was set at 200 and the weight decay at 0.01. All four models achieved an AUC of 0.95 (Table~\ref{tab:transfer_eff}). Taking a model average improved the AUC to 0.98 which corresponded to sensitivity of 86.7\% and specificity of 96.1\% (Fig.~\ref{fig:roc_curve}b).

\begin{table}
\centering
\caption{Transfer learning efficiency with different training set sizes. Shown are per-image validation AUCs.}
\label{tab:transfer_eff}
\begin{tabular}{@{}llllll@{}}
\toprule
\textbf{\#Patients} & \textbf{\#Images} & \textbf{Resnet-Resnet} & \textbf{Resnet-VGG} & \textbf{VGG-VGG} & \textbf{VGG-Resnet} \\ \midrule
20 & 79 & 0.92 & 0.88 & 0.87 & 0.89 \\
30 & 117 & 0.93 & 0.94 & 0.93 & 0.90 \\
40 & 159 & 0.93 & 0.95 & 0.93 & 0.93 \\
50 & 199 & 0.94 & 0.95 & 0.94 & 0.93 \\
60 & 239 & 0.95 & 0.95 & 0.95 & 0.94 \\
72 (All) & 280 & 0.95 & 0.95 & 0.95 & 0.95 \\ \bottomrule
\end{tabular}
\end{table}

We then sought to determine the minimum amount of data required to fine-tune a whole image classifier to a satisfactory level of performance, thereby minimizing the resource intensive process of obtaining labels. Training subsets with 20, 30, 40, 50 and 60 patients were sampled for fine-tuning a model and evaluating the model's performance on the same validation set (Table~\ref{tab:transfer_eff}). With as few as 20 patients or 79 images, the four models were already achieving AUCs that varied between 0.87-0.92. The AUCs quickly approached the maximum as we increased the training subset size. We hypothesize that the "hard" part of learning is to recognize the shapes and textures of the benign and malignant ROIs and normal tissues, while the "easy" part is to adjust to different intensity profiles on independent datasets. Importantly, these results clearly demonstrate that the end-to-end training approach can be successfully used to fine-tune a whole image classifier using additional small training sets with image-level labels, greatly reducing the burden of training set construction.

\section{Discussion}
Our findings show that accurate classification of screening mammograms can be achieved with a deep learning model trained in an end-to-end fashion that relies on clinical ROI annotations only in the initial stage. Once the whole image classifier is built, it can be fine-tuned using additional datasets that lack ROI annotations, even if the pixel intensity distributions differ as is often the case for datasets assembled from heterogeneous mammography platforms. Our findings show that deep learning algorithms can outperform current commercial CAD systems which have been reported to attain an average AUC of 0.72 \cite{cole_impact_2014}. Our all convolutional networks trained using an end-to-end approach also have highly competitive performance and are more generalizable across different mammography platforms compared to previous deep learning methods that have achieved AUCs in the range of 0.65-0.97 on the DDSM and INbreast databases, as well as in-house datasets \cite{burt_deep_2018}.

Two recent studies \cite{zhu_deep_2017,choukroun_mammogram_2017} developed deep learning based methods for breast cancer classification using film and digital mammograms, which are end-to-end trainable. Each study uses multi-instance learning (MIL) and modifies the whole image classifiers' cost functions to satisfy the MIL criterion. In contrast to our approach, neither study utilizes ROI annotations to train the patch classifiers first and the AUCs are lower than reported in this study. We found that the quality of the patch classifiers is critical to the accuracy of the whole image classifiers. This was supported by two lines of evidence: First, the whole image classifier based on the S10 patch set performed far better than the one based on the S1 patch set (Table~\ref{tab:resnet_auc}). Second, it took much longer for the VGG16-based whole image classifiers to achieve the same performance as the Resnet50-based classifiers (Tables \ref{tab:resnet_auc} \& \ref{tab:vgg_auc}) because VGG16 was less accurate than Resnet50 in patch classification (Table~\ref{tab:patch_acc}).

We also found that sampling more patches to include the ROIs' neighboring and background regions improved the whole image classification. However, the computational burden increases linearly with the number of patches sampled and the performance gain may quickly diminish. Thus, further research is needed to investigate how to sample local patches more efficiently—--perhaps by focusing on the patches that are likely to be misclassified—--to help overcome the computational burden of training more accurate patch classifiers.

Although the performance between VGG-based and Resnet-based whole image classifiers was comparable, the VGG-based classifiers tended to overfit and required longer training. On the other hand, the VGG16 (without the two FC layers), with 15 million weight parameters, is a much smaller network than the Resnet50, with 24 million weight parameters. Having fewer parameters reduces memory usage and training time per epoch, which are desirable when computational resources are limited. The Resnet is a more recently developed deep learning method, which is enhanced by shortcuts and batch normalization, both techniques may help the network train faster and generalize better. The same techniques can be used on the VGG-based networks as well, which may provide future improvement for the VGG-based classifiers.

In this study, the mammograms were down-sized to fit into the available GPU (8GB). As more GPU memory becomes available, future studies will be able to train models using larger image sizes, if not retain the native image resolution without the need for downsizing. Retaining the full resolution of modern digital mammography images will provide finer details of the ROIs and would likely improve performance. 

In conclusion, our study demonstrates that deep learning models trained in an end-to-end fashion are highly accurate and can be readily transferred across heterogeneous mammography platforms. Thus, deep learning methods have enormous potential to improve the accuracy of breast cancer detection on digital screening mammograms as the available training datasets expand. Our approach may assist in the development of superior CAD systems that can increase the benefit and reduce the harm of screening mammography and may have applications in other medical imaging problems where ROI annotations are scarce.

\section*{Acknowledgments}
This work was partially supported by the Friedman Brain Institute and the Tisch Cancer Institute (NIH P30 CA196521). The funders had no role in study design, data collection and analysis, decision to publish, or preparation of the manuscript. This work was supported in part through the computational resources and staff expertise provided by the Department of Scientific Computing at the Icahn School of Medicine at Mount Sinai. We would like to thank Gustavo Carneiro, Gabriel Maicas, Daniel Rubin and Meng Cao for providing comments on the manuscript, and Quan Chen for discussion on the use of the INbreast data.

\section*{Author contributions statement}
L.S. developed the algorithm, conceived and conducted the experiments and analyzed the results. L.R.M., J.R., R.M. and W.S. analyzed the results. E.F. helped with the computational resources. All authors contributed to writing and reviewing the manuscript.

\section*{Computational environment}
All experiments in this study were carried out on a Linux workstation equipped with an NVIDIA 8GB Quadro M4000 GPU card.

\printbibliography

\end{document}